%% file: iclr2021_conference.tex
\documentclass{article} 
\usepackage{iclr2021_conference,times}
\usepackage[hidelinks,colorlinks=true, linkcolor=black,urlcolor=blue,citecolor=blue]{hyperref}
\input{math_commands.tex}

\usepackage{CJKutf8}
\usepackage[utf8]{inputenc} 
\usepackage[T1]{fontenc}    
\usepackage{hyperref}       
\usepackage{url}            
\usepackage{booktabs}       
\usepackage{amsfonts}       
\usepackage{nicefrac}       
\usepackage{microtype}      
\usepackage{lipsum}
\usepackage{ragged2e,array,booktabs}
\usepackage{fancyhdr}       
\usepackage{graphicx}       
\graphicspath{{media/}}     
\usepackage{caption}
\usepackage{titlecaps}
\usepackage{amsmath} 
\usepackage{amssymb} 
\usepackage{relsize}
\usepackage{tablefootnote}
\usepackage{threeparttable}
\usepackage{multirow}
\usepackage{enumitem}
\usepackage{makecell}
\usepackage{etoolbox}

\appto\TPTnoteSettings{\footnotesize}
\newcolumntype{P}[1]{>{\centering\arraybackslash}p{#1}}

\title{Bailong: Bilingual Transfer Learning based on QLoRA and Zip-tie Embedding}

\author{Lung-Chuan Chen \\ INNOLUX \\ \texttt{blaze7451@gmail.com} \And
        Zong-Ru Li\\ INNOLUX \\ \texttt{z7676x52678d@gmail.com}}

\graphicspath{ {./Figures/} }
\iclrfinalcopy 
\begin{document}

\begin{CJK*}{UTF8}{bsmi}
\maketitle

\begin{abstract}
Large language models (LLMs) have demonstrated exceptional performance in various NLP applications. However, the majority of existing open-source LLMs are pre-trained primarily on English data and little part of other languages. This deficiency in multilingual training data results in suboptimal performance when applied to languages with fewer available resources. Furthermore, enhancing the performance of Large Language Models (LLMs) on low-resource languages by full-parameter fine-tuning with additional data requires substantial computational resources, posing computational barriers for research organizations and individual researchers. Consequently, several techniques such as parameter-efficient tuning and advanced embedding initialization have been proposed to address these challenges. In this work, we combine them to effectively facilitate cross-lingual transfer on English-dominated open-source LLM. Specifically, to effectively enhance the model's proficiency in Traditional Chinese, we conduct secondary pre-training on Llama 2 7B with Traditional Chinese data by leveraging QLoRA and our proposed zip-tie embedding initialization.  The resulting model called \textbf{Bailong}, which stands for \textbf{B}ilingual tr\textbf{A}nsfer learn\textbf{I}ng based on q\textbf{LO}ra and zip-tie embeddi\textbf{NG}. Additionally, we present Bailong-instruct 7B, a fine-tuned version of Bailong 7B optimized for multi-turn dialogue scenarios. Recognizing the inadequacy of benchmark datasets in Traditional Chinese, we further introduce Bailong-bench to assess the alignment of models with human preferences and their capability to follow instructions in both Traditional Chinese and English tasks. In our evaluation, Bailong-instruct 7B exhibits competitive performance on Bailong-bench and other benchmark datasets when compared to other open-source models of similar or even larger parameter sizes. Bailong-instruct 7B\footnote{\url{https://huggingface.co/INX-TEXT/Bailong-instruct-7B}} and Bailong-bench\footnote{\url{https://huggingface.co/datasets/INX-TEXT/Bailong-bench}} are publicly available with the aim of empowering the community to build upon our efforts and actively participate in the development of LLMs. 
\end{abstract}

\section{Introduction}
With rapid development in recent years, large language models (LLMs) have been widely utilized across various practical domains due to their remarkable abilities in contextual comprehension, reasoning, and text generation. One of the most notable applications is the use in chatbot implementations, which allows humans to interact directly with them through an intuitive user interface. These chat models perform a range of natural language processing (NLP) tasks by analyzing natural language inputs, showcasing exceptional performance in tasks like sentiment analysis, summarization, translation, and so forth. Additionally, LLMs have recently been enhanced with multimodal capabilities when intergated with visual encoders \citep{dai2023instructblip, bai2023qwen, liu2023improved}. This development has led to the recognition of LLMs as potential candidates for artificial general intelligence (AGI) or for applications in more advanced domains, prompting thorough and proactive research efforts in both academia and industry.

Despite the great potential of these models, in numerous situations, they are closed-source and only accessible with proprietary APIs, restricting the progress of research and further development. In contrast to proprietary models proffered by corporate entities such as ChatGPT\citep{Citechatgpt2022}, Gemini \citep{team2023gemini}, and GPT4 \citep{achiam2023gpt}, the more transparent and adaptable open-source large-scale language models like Llama 2 \citep{touvron2023llama}, Gemma \citep{team2024gemma}, and Mistral series \citep{jiang2023mistral} are released to facilitate research across various dimensions, thereby empowering the industry or research community to applications or develop on top of these models.

However, in spite of the considerable efforts and benefits that open-source large language models bring to the NLP community, practical deployment in real-world scenarios still remains challenging. The training corpora used during the pre-training phase of such models are primarily composed of English, with significantly lower proportions of training data in other languages. For instance, in the case of Llama 2, training data for languages other than English and unknown languages constitutes approximately only 1.92\% of the total corpus. In contrast to the outstanding performance of the model on English tasks, its performance is suboptimal on other low-resource languages such as Traditional Chinese and Thai. Furthermore, enhancing the performance of Large Language Models (LLMs) on low-resource languages by full-parameter fine-tuning with additional data requires substantial computational resources, posing computational barriers for research organizations and individual researchers.

To address these issues, we leverage QLoRA \citep{dettmers2023qlora} and our proposed zip-tie embedding to effectively and efficiently implement cross-lingual transfer learning on large language model. The resulting model is called \textbf{Bailong}, which stands for \textbf{B}ilingual tr\textbf{A}nsfer learn\textbf{I}ng based on q\textbf{LO}ra and zip-tie embeddi\textbf{NG}. In order to verify the effectiveness of our training method, we apply it to enhance the model's proficiency in Traditional Chinese. Firstly, inspired by \citep{cui2023efficient}, we initially expand the vocabulary of Llama 2 7B with additional 27,241 Traditional Chinese tokens to improve the encoding and decoding efficiency of the model in Traditional Chinese. Leveraging the QLoRA technique, we apply LoRA layers not only to the MLP and attention layers but also adjacent to embedding layer and language model head for secondary pre-training of the model. This further reduces the number of parameters that needed during the secondary pre-training phase. Additionally, we introduce a novel initial embedding strategy—zip-tie embedding. The use of zip-tie embedding helps lower the initial loss value during model training and effectively reduces the training steps. This introduces a potential research direction, allowing developers to efficiently train models by optimizing the strategy for initial embedding. The resulting model is called Bailong 7B. We also trained Bailong-instruct 7B, a instruction fine-tuned version of Bailong 7B, strengthening its ability to generate accurate responses aligned with human-like instructions for multi-turn dialogue use cases. Finally, to evaluate the performance of the model in real-world applications and tasks involving Traditional Chinese and English instructions, we introduce Bailong-bench, a benchmark dataset designed for assessing the model's performance in real-world scenarios including creative writing, proofreading, machine translation, summarization and so on. Besides evaluating on Bailong-bench, we also assess the performance of Bailong-instruct 7B on other Traditional Chinese datasets, thereby demonstrating the effectiveness of our proposed method.

In a nutshell, the contributions of our work can be organized as follows:
\begin{itemize}
\item We extend the vocabulary of Llama 2-7B with additional 27,241 Traditional Chinese tokens. Leveraging QLoRA technique, we deployed LoRA layers across all MLP, attention layers, language model head, and the embedding layer. This substantial reduction in the number of parameters required during fine-tuning in the secondary pre-training phase resulted in the model we refer to as Bailong-7B.
\item We introduce zip-tie embedding, a novel embedding initialization method which helps to lower the initial loss value during cross-lingual transfer and save the training steps.
\item We further leverage QLoRA for instruction fine-tuning on Bailong 7B. The resulting model trained in this stage is named as Bailong-instruct 7B.
\item We introduce Bailong-bench, a benchmark dataset comprising 140 questions. Through Bailong-bench, people can assess the consistency of models with human preferences in open-ended tasks in both Traditional Chinese and English.
\end{itemize}

\section{Related Work}
\paragraph{Traditional Chinese LLMs}
Continuously pre-training the open-source LLMs on a low-resource target language, whether accompanied by the inclusion of supplementary English data or not, has shown a noteworthy improvement in model performance on low-resource language tasks \citep{pires2023sabi, pipatanakul2023typhoon, CiteOpenhathi2023}. In order to boost the model's proficiency in Traditional Chinese, \citep{ennen2023extending} collect and preprocess the publicly available and authorized private datasets to implement continually pre-training on BLOOM series with additional billions of tokens \citep{workshop2022bloom}. The resulting model series, dubbed BLOOM-zh, surpasses its predecessor on most Traditional Chinese benchmarks. Taiwan-LLM \citep{lin2023taiwan} continues pre-training Llama 2 7B and Llama 2 13B on Traditional Chinese data from various sources and performs supervised fine-tuning and feedback supervised fine-tuning to strengthen the instruction-following ability of models. The objective of Taiwan-LLM is to develop language models that not only excel in Traditional Chinese NLP tasks but also demonstrate a deep understanding of Taiwanese culture. Both of BLOOM-zh and Taiwan-LLM series are trained using a full-parameter training method, which necessitates considerable computational resources and is time-consuming to reach optimal performance. Compared to these works, our work focuses on efficiently and effectively enhancing model performance on Traditional Chinese tasks through parameter-efficient tuning. 

\paragraph{Evaluation of Traditional Chinese LLMs}
In the context of Traditional Chinese, the main challenges for building LLMs include not only collecting a sufficient amount of Traditional Chinese data of high quality, but also the lack of comprehensive benchmarks to assess the capability of LLMs across various domains. The existing Traditional Chinese benchmarks are mostly curated to evaluate core performance of language model solely on a limited set of tasks. For instance, DRCD (Delta Reading Comprehension Dataset) \citep{shao2018drcd} is a reading comprehension dataset containing more than 30000 data retrieved from Wikipedia; FGC (Formosa Grand Challenge) \citep{CiteFGC2020} is another reading comprehension based question answering dataset provided by Taiwan government; TTQA (Taiwanese Trivia Question Answering) \citep{ennen2023extending}, comprising 64 paragraphs data extracted from Wikipedia, is an evaluation dataset for testing commonsense knowledge ability of models on Taiwan topics; TMMLU (Taiwan Massive Multitask Language Understanding) \citep{hsu2023advancing} is a massive world knowledge evaluation dataset encompassing examination questions across 55 subjects. While these benchmarks have exhibited a degree of usefulness on assessing model performance, their limitations appear when measuring model alignment with human preference on open-ended tasks. This deficiency results in a notable disparity between the model's scoring-based performance and its practical applicability in real-world scenarios. Different from these works, our proposed Bailong-bench is designed to evaluate LLM alignment with human preferences such as instruction-following and multi-turn dialogue abilities on open-ended tasks. 

\paragraph{Parameter-efficient cross-lingual transfer learning}
Parameter-efficient fine tuning (PEFT) methods refer to injecting trainable lightweight modules into the layers of model architecture and freezing parameters of the pretrained model during fine-tuning. These approaches have been shown to allow for model adaptation in a resource-constrained setting while yielding competitive outcomes comparable to a fully fine-tuned model. In language adaptation scenario, numerous variants of Adapter \citep{DBLP:journals/corr/abs-1902-00751} method are leveraged to perform cross-lingual transfer in early studies. As one example of them, MAD-X \citep{pfeiffer2020mad} uses language adapters and invertible adapters to learn each language and stacks task adapters on top of language adapters when training on a downstream tasks. After training, MAD-X implements zero-shot cross-lingual transfer by simply replacing source language adapters with target language counterpart. Instead of training language adapters for each individual languages, BAD-X \citep{parovic2022bad} leverages bilingual language adapters to boost the performance of downstream tasks on a specific source-target transfer regime. For improving the robustness of adapters to unsupported language without training additional adapters, \citep{wang2021efficient} ensemble the existing language adapters by entropy of minimized ensembling of adapters (EMEA) method.

Even though the adapter-based methods have shown their great effectiveness and efficiency on language adaption, they induce additional inference latency during inference due to their sequentially inserting design. In contrast, Low-Rank Adaptation (LoRA) \citep{hu2021lora} method inserts low-rank matrices besides the linear layers of transformer architecture in parallel, enabling the merge between LoRA matrices and frozen weights, hence introducing no inference latency during inference. Concretely, consider a linear layer with pre-trained weight matrix $\text{W}\in\mathbb{R}^{d\times k}$, where $d$ denotes the output dimension and $k$ denotes the input dimension, LoRA represents the weight change matrix $\Delta W$ as the product of two low-rank matrices $\text{L}_{1}\in \mathbb{R}^{d\times r}$ and $\text{L}_{2}\in \mathbb{R}^{r\times k}$, where $\text{rank}\, r \ll \min(d, k)$. Given an input $\text{X}\in\mathbb{R}^{k\times o}$ and an output $\text{Y}\in\mathbb{R}^{d\times o}$, the forward pass is expressed as:
\begin{equation} \label{eqn:loraformula}
    \text{Y} = \text{W}\text{X} + \text{L}_{1}\text{L}_{2}\text{X},
\end{equation}
where $W$ is frozen during training. Closet to our work, \citep{cui2023efficient} incorporate LoRA matrices into the weights of attention modules and MLP layers during pre-training and fine-tuning stages. By training the embedding, LM head, and newly added LoRA parameters, they efficiently improve Chinese understanding and enhance the ability of following human instructions of models. To further reduce the numbers of trainable parameters, instead of training embedding layer and LM head directly, we add LoRA matrices beside them and freeze their parameters when training.

\section{Continual Pre-training}

\subsection{Method}
\paragraph{Vocabulary extension}

\begin{table}
\centering
\resizebox{\textwidth}{!}{%
\begin{tabular}{l>{\RaggedRight}p{10cm}l}
\hline
Type & Content & Length\\
\hline
Original sentence & 通過創建新理論與發展新科技，物理學對於人類文明有極為顯著的貢獻。 & 32 \\
\hline
Llama 2 tokenizer & '▁', '通', '<0xE9>', '<0x81>', '<0x8E>', '<0xE5>', '<0x89>', '<0xB5>', '建', '新', '理', '論', '<0xE8>', '<0x88>', '<0x87>', '<0xE7>', '<0x99>', '<0xBC>', '展', '新', '科', '技', '，', '物', '理', '學', '<0xE5>', '<0xB0>', '<0x8D>', '<0xE6>', '<0x96>', '<0xBC>', '人', '<0xE9>', '<0xA1>', '<0x9E>', '文', '明', '有', '<0xE6>', '<0xA5>', '<0xB5>', '<0xE7>', '<0x82>', '<0xBA>', '<0xE9>', '<0xA1>', '<0xAF>', '<0xE8>', '<0x91>', '<0x97>', '的', '<0xE8>', '<0xB2>', '<0xA2>', '<0xE7>', '<0x8D>', '<0xBB>', '。' & 59  \\
\hline
Bailong tokenizer & '▁通', '過', '創', '建', '新', '理論', '與發展', '新', '科技', '，', '物理學', '對於', '人類', '文明', '有', '極為', '顯著的', '貢獻', '。' & 19 \\
\hline
\end{tabular}}
\caption{Comparison between Llama 2 tokenizer and Bailong tokenizer.}
\label{table:comparison between llama and bailong}
\end{table}

Original Llama 2 tokenizer was trained by utilizing BBPE \citep{wang2020neural} technique to avoid Out-of-Vocabulary (OOV) problem and maintain compactness concurrently. Within this framework, characters of character-rich languages such as Chinese and Japanese are typically tokenized into multiple byte-level subwords and thus poses significant constraints on the effective sequence lengths of these languages during the process of language modeling. To tackle this issue, we implement byte-pair encoding (BPE) algorithm \citep{sennrich2015neural} to train additional tokenizer on Traditional Chinese Wikipedia data. The resulting tokenizer has a vocabulary size of 30000. Subsequently, we merge our Traditional Chinese tokenizer into Llama 2's tokenizer and remove the tokens shared between them. The merged tokenizer, dubbed Bailong tokenizer, has a vocabulary size of 59271 eventually. As Table \ref{table:comparison between llama and bailong} shown, when applied to the same Traditional Chinese sequence, Bailong tokenizer uses only one-third of the tokens compared to Llama 2 tokenizer for tokenization, allowing model to handle longer sequence during training with the fixed computational budget. To measure encoding efficiency of tokenizer on Traditional Chinese, we define efficiency as the number of standard tokens with respect to the number of tokens generated from model tokenizer on the same treebank sentence stored in Chinese GSD subset\footnote{\url{https://github.com/UniversalDependencies/UD_Chinese-GSD/tree/master}} of Universal Dependencies (UD) \citep{nivre2020universal}, i.e.,
\begin{equation}
    \text{Efficiency} = \frac{\#\text{token}_{\text{standard}}}{\#\text{token}_{\text{model}}},
\end{equation}
where $\#\text{token}_{\text{standard}}$ is the number of gold standard tokens defined by UD authors and $\#\text{token}_{\text{model}}$ is the number of tokens generated from model tokenizer. Since there are totally 4997 treebank sentences preserved in Chinese GSD subset of UD, we calculate the total number of tokens for all sentences and divide it by the overall number of tokens generated by the model after tokenizing all sentences. We report the efficiencies of Bailong tokenizer and other tokenizers in Table \ref{table:Efficiency of tokenization}. As expected, Bailong tokenizer demonstrates superior segmentation efficiency in Traditional Chinese compared to character-level segmentation. Furthermore, it outperforms other tokenizers without an extended vocabulary, albeit with a slight lag behind the segmentation approach employed by Jieba, a specialized module designed for Chinese segmentation. 

\begin{table}[h]
\centering
\begin{threeparttable}
\begin{tabular}{l|cccccc}
\hline
Tokenizer & Character & Taiwan-LLM & GPT-4 & Gemma & Bailong & \text{Jieba\tnote{1}}\\
\hline
Efficiency & 0.63 & 0.39 & 0.44 & 0.80 & 1.01 & 1.08\\
\hline
\end{tabular}
\begin{tablenotes}
     \item[1] \url{https://github.com/fxsjy/jieba}
\end{tablenotes}
\caption{Efficiency of tokenization on Traditional Chinese text in Chinese GSD subset of UD.}
\label{table:Efficiency of tokenization}
\end{threeparttable}
\end{table}

After merging the tokenizer, we resize the embedding layer and language model head of Llama 2 from shape $32, 000\times D$ to $59271\times D$, where $D$ denotes dimension of the embedding layer. With the procedure of resizing, the appended vectors related to newly introduced tokens are initialized from a default normal distribution while the original vectors remain unaltered.

\begin{figure}
  \centering
  \includegraphics[width=\linewidth]{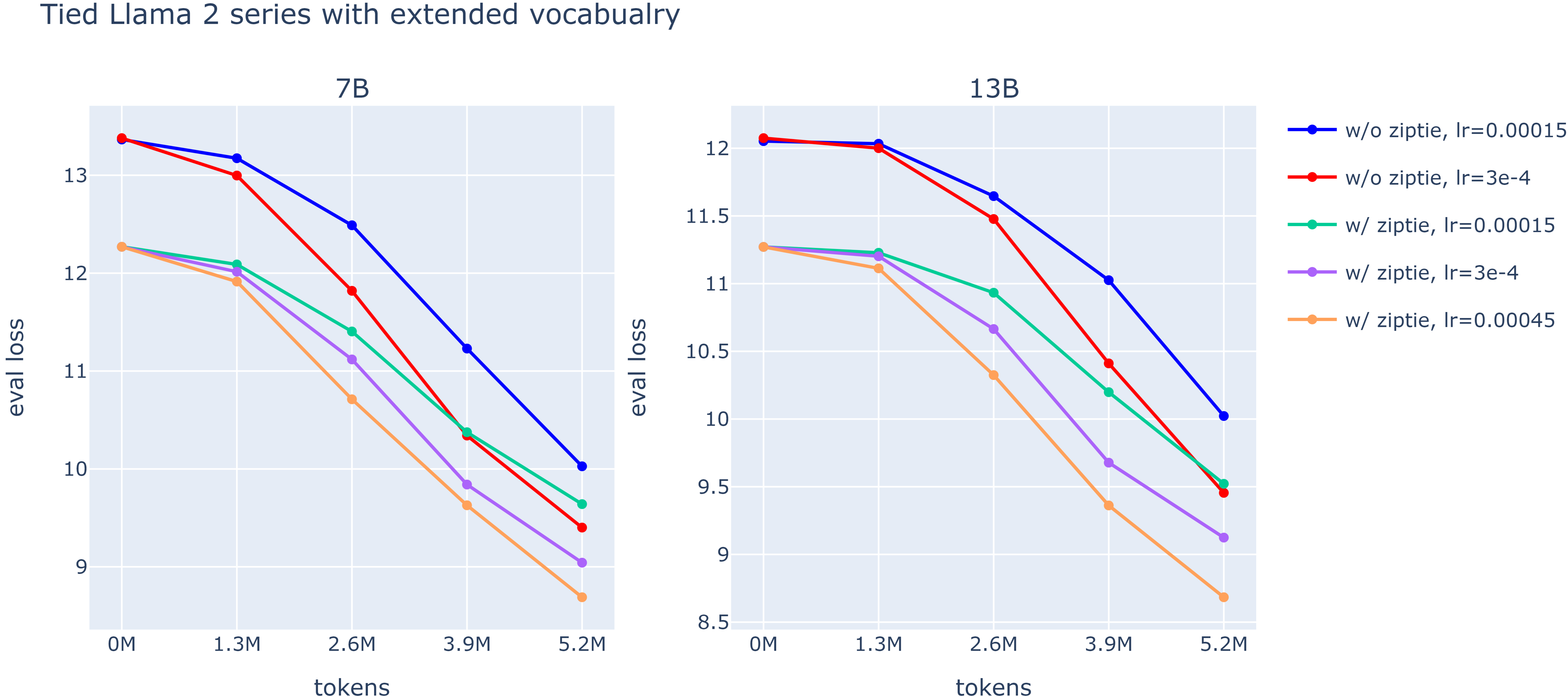}
  \caption{Losses over training tokens from training Llama 2 7B and Llama 2 13B on Trditional Chinese subset of Wikipedia. The evaluation is performed every 1.3 million trained tokens.}
  \label{fig:experimental result of zip-tie embedding}
\end{figure}

\paragraph{Zip-tie embedding initialization}
Despite the observation that solely adding embedding vectors initially has a negative impact on model, they have been shown to eventually enhance model performance through continual pre-training \citep{si2023empirical}. For efficiently adapting language model to new language, one of the primary challenge lies in aligning the embeddings of the new target language with the original embedding of language models by leveraging the knowledge embedded within it \citep{pfeiffer2020unks}. Prior study utilizes embeddings of lexical overlapping tokens between source language and target language to initialize target embeddibngs of tokens which are not in the overlapping vocabulary \citep{ostendorff2023efficient}. Consider the original English-centric multilingual embeddings $\text{\textbf{E}}_{s}\in\mathbb{R}^{\lvert V\rvert\times D}$ which projects each token $v_{s}$ in vocabulary $V$ to the vector representation $\bm{v}_{s}\in\mathbb{R}^{D}$, the extended vocabulary is expressed as $V' = V \cup V_{add}$, where $V_{add} = V' \setminus V$ is the newly added Traditional Chinese vocabulary in our case. Each Traditional Chinese token $v_{t}$ in $V_{add}$ can be tokenized into a set of byte-level tokens $V_{byte, t}=\{v_{t1},...,v_{tn}\}\in V$. The core idea of zip-tie embedding arises from the fact that each representation of newly added token within the expanded vocabulary corresponds to multiple byte-level tokens in the original model. By leveraging this relationship, we hypothesis that the embeddings of these additional tokens can be initialized by combining the embeddings of their respective byte-level counterparts. To be specific, we initialize the embedding $\bm{v}_{t}$ as:
\begin{equation}\label{eqn:zip-tie embedding}
    \bm{v}_{t} = \sum_{v_{ti}\in V_{byte, t}}\alpha(v_{t}, v_{ti})\bm{v}_{ti},
\end{equation}
where $\alpha$ is the weight function of the embeddings of newly added token $\bm{v}_{t}$ and the embedding of its related byte-level token $\bm{v}_{ti}$. The goal is then to find a appropriate $\alpha$ to effectively transfer the inherent lexical knowledge from $\bm{v}_{ti}$ to $\bm{v}_{t}$. 
One possible method is to compute $\alpha$ as normalized cosine similarity of embeddings existing in the smaller model \citep{ostendorff2023efficient}. However, this method is infeasible if there is no smaller model existing. In contrast to them, \citep{tran2020english} computes the weight function as the product of static word embdddings instead of contextualized counterparts. We leave the exploration of advanced weight function to future study. In this work, we simply defined $\bm{v}_{t}$ as the averaged sum of embeddings of the byte-level tokens in $V_{byte, t}$. We refer this method as zip-tie embedding initialization, and the resulting embedding $\bm{v}_{t}$ is called zip-tie embedding. To empirically validate the usefulness of our method, we tie the embedding layers and LM heads of vocabulary-extended Llama 2 models and apply zip-tie embedding initialization to them. Figure \ref{fig:experimental result of zip-tie embedding} shows that despite its simple design, vocabulary-extended moedls trained with zip-tie embedding initialization have lower evaluation loss values compared to the ones with random initialization. This result showcases a promising direction to utilize more advanced embedding initialization strategies to facilitate cross-lingual transfer of monolingual LLMs.

\paragraph{Parameter-efficient tuning with QLoRA}
The implementation of LLMs demands substantial computational resources and memory, thereby posing a formidable obstacle to the applicability and scalability. Quantization techniques involve the conversion of model weights from high-bit format to low-bit counterpart to reduce memory requirement while incurring performance degradation at times. On top of LoRA, QLoRA introduces multiple novel features to save memory usage during training without losing performance, including 4-bit NormalFloat (NF4) quantization, Double Quantization (DQ) method, and paged optimizers \citep{dettmers2023qlora}. 

To be specific, with the implementation of QLoRA, \eqref{eqn:loraformula} is rewritten as:
\begin{equation}
    \text{Y}^{\text{BF16}} = \text{doubleDequant}(\text{c}_{1}^{\text{FP32}}, \text{c}_{2}^{\text{FP8}},W^{\text{NF4}})\text{X}^{\text{BF16}} +  \text{L}_{1}^{\text{BF16}}\text{L}_{2}^{\text{BF16}}\text{X}^{\text{BF16}},
\end{equation}
where $\text{c}_{1}$ and $\text{c}_{2}$ are the second level and the first level of quantization constants respectively, the pre-trained weight matrix $W$ is now preserved in NF4 format, and the $\text{doubleDequant}(\cdot)$ is defined as:
\begin{equation} 
    \text{doubleDequant}(\text{c}_{1}^{\text{FP32}}, \text{c}_{2}^{\text{FP8}},W^{\text{NF4}}) = \text{dequant}(\text{dequant}(\text{c}_{1}^{\text{FP32}},  \text{c}_{2}^{\text{FP8}}),W^{\text{NF4}}) = W^{\text{BF16}},
\end{equation}
where $\text{dequant}(\cdot)$ is the dequantization:
\begin{equation}
    \text{dequant}(\text{c}_{1}^{\text{FP32}}, \text{c}_{2}^{\text{FP8}}) = \frac{\text{c}_{2}^{\text{FP8}}}{\text{c}_{1}^{\text{FP32}}} = \text{c}_{2}^{\text{FP32}}.
\end{equation}

\subsection{Pre-training Data}
Curation of training dataset is crucial and indispensable in the model development pipeline. Prior studies \citep{conneau2019unsupervised, penedo2023refinedweb} have demonstrated that models trained on large scale web data with careful filtering and deduplication can lead to powerful or even state-of-the-art performance. However, owing to the limited availability of data for low-resource languages, the creation of a genuinely low-resource language training dataset for LLMs with billions or trillions of parameters becomes prohibitively challenging. Achieving a dataset that exclusively consists of web content while concurrently adhering to the scaling law between training tokens and model parameters can be an insurmountable task at times. Notably, to reach the optimized result, the model capacity and the number of training tokens should be scaled equally \citep{hoffmann2022training}. As a result, to scale our training dataset as large as possible, we opt to build our training dataset by collecting web data and high quality curated corpora including books, Wikipedia data, and human conversations.

\subsubsection{Data Collection}
Our training dataset is primarily composed of Traditional Chinese data with a minor portion of English content. For the English language data, we utilize small subsample of original RefinedWeb \citep{penedo2023refinedweb} without additional processing. For the Traditional Chinese language data, in order to scale up the dataset to desirable size, we collect publicly accessible datasets in both Simplified Chinese and Traditional Chinese initially. Specifically, our dataset contains few curated corpora containing Simplified Chinese and Traditional Chinese Wikipedia\footnote{\url{https://dumps.wikimedia.org/zhwiki/}}, YeungNLPfirefly-train-1.1M\footnote{\url{https://huggingface.co/datasets/YeungNLP/firefly-train-1.1M}}, and Zhihu-KOL\footnote{\url{https://huggingface.co/datasets/wangrui6/Zhihu-KOL}}.

Futhermore, we apply trafilatura library \citep{barbaresi-2021-trafilatura} to extract human curated contents from meticulously chosen web pages across diverse categories, such as blog posts, news, and column articles. Throughout this procedure, we only crawled the web pages which are originally edit in Traditional Chinese to ensure a certain proportion of Traditional Chinese data in our raw dataset. We gather the publicly accessible books written in Simplified Chinese and Traditional Chinese as well. The resulting two datasets, dubbed WebDocs and Books, contain approximately one billion and 100 million tokens respectively. As the remaining web data portion, our dataset includes Chinese-language subset of OSCAR 23.01\footnote{\url{https://oscar-project.github.io/documentation/versions/oscar-2301/}} and part of Common Crawl\footnote{\url{https://commoncrawl.org/overview}} dataset which is composed of Traditional Chinese and Simplified Chinese documents. In the end of this stage, without any filtering or processing, our whole initial training dataset includes around 400 billion tokens.

\subsubsection{Filtering and Converting}
This section describes several preprocessing methods we perform to enhance the quality of the corpora. All of these methods are applied across the whole training dataset.  

\paragraph{Simplified Chinese conversion} We use OpenCC\footnote{\url{https://github.com/BYVoid/OpenCC}} to convert Simplified Chinese data into Traditional Chinese. During the procedure, some of the idioms are transformed into commonly used Taiwanese terms as well. Moreover, to better match the preferences of Traditional Chinese users, we transform the half-width punctuation marks into their full-width counterparts.

\paragraph{Banned words filtering} To wipe out explicit contents such as sex, violence, and profanity from our corpora, we construct a banned word list to filter the data that contains words in the list. Throughout the procedure, the majority of the corpora only contain a small portion of data with banned words. Nonetheless, we observe that approximately 90\% of data in the "zh" subset of OSCAR 23.01 dataset has sexual content and is therefore discarded.

\paragraph{Removal of special symbols, URLs, emojis, and emails} Sentences that contains special symbols, URLs, emojis, or emails are excluded due to the potentiality of their impact on model's performance and issue of privacy. We also remove the data which contains less than 10 tokens counted by Bailong tokenizer.

\subsubsection{Deduplication}
Several works in CV and NLP have shown that training data went through thoughtful data selection not only accelerates training but also leads to superior performance compared to using uncurated data \citep{xie2023data, sorscher2022beyond, coleman2019selection, zhao2023dataset}. For text data, early works leverage trained binary classifier to filter web text which is predicted to be lower quality \citep{brown2020language, gao2020pile}. Recent studies indicate increasing focus on advanced data-selecting strategies based on various deduplication algorithms given that duplicated data can lead to performance degradation and induce model to shift from generalization to memorization \citep{hernandez2022scaling}. These deduplication methods can be performed with the use of exact matching and approximate matching, depending on the type of the duplicates. For exact matching, suffix array \citep{manber1993suffix} and longest common substring \citep{abdi2023longest} are commonly performed to search the verbatim substrings between documents. For fuzzy matching, MinHash\citep{broder1997resemblance} is the most common algorithm to find the similar documents in the text sets. Besides these methods, other works \citep{abbas2023semdedup, tirumala2023d4} have demonstrated that removing semantic duplicates could substantially boost training and achieve better result than random selection.

In this work, to ensure the diversity of training data and avoid semantic and textual repetition, after the procedures of filtering and converting, we implement both traditional locality-sensitive hashing algorithm and semantic deduplication. We find that they are memory efficient and can effectively remove the duplicates as we need.

\paragraph{Fuzzy Deduplication.} We apply MinHash \citep{broder1997resemblance} in conjunction with Locality-Sensitive Hashing (LSH) to estimate the Jaccard similarity of each document pair and remove one of the documents in the pair if the documents are identified as duplicates. Concretely, we initially tokenize the documents in the dataset via our Bailong tokenizer to represent each document as a set of 1-grams. Implementing MinHash then constructs hash signatures for each set with the use of numerous permutation functions. Following this step, we proceed to create LSH table using a collection of hash functions and map MinHash signatures to hash buckets. Therefore, we can query the potential matches by mapping query documents to the hash buckets and retrieve documents which fall into the same bucket as the query document. For document pairs identified as potential matches, we compute their MinHash-based Jaccard similarity and identified them as duplicates if the similarity score surpasses the default threshold. In our implementation, we set the number of permutation functions to 256 and the similarity threshold to 0.8 as this setting achieves a balance between performance and computational budget.

\paragraph{Semantic deduplication.} We perform SemDeDup \citep{abbas2023semdedup} with a little bit modification to remove the semantic duplicates in each corpus of the entire training dataset. Instead of the method in \citep{abbas2023semdedup} in which the authors proposed to leveraging embedding vectors from pretrained language models, we harness pretrained multilingual sentence transformer models \citep{reimers2019sentence} within widely recognized SentenceTransformers\footnote{\url{https://www.sbert.net/docs/pretrained_models.html}} library for data embedding. We apply "paraphrase-multilingual-MiniLM-L12-v2" and "paraphrase-multilingual-mpnet-base-v2", two multilingual sentence transformers pretrained on data for more than 50 languages, to implement the document embedding in each dataset. We observe that they both perform better on Traditional Chinese language embedding compared to OPT model. We also find that the performance difference between "paraphrase-multilingual-MiniLM-L12-v2" and "paraphrase-multilingual-mpnet-base-v2" are small in most of the case. Considering about the computational budget, we decide to utilize "paraphrase-multilingual-MiniLM-L12-v2" to implement embedding for all corpora. Data points after embedding are then clustered into k clusters via k-means and the point pairs within the same cluster are considered as semantic duplicates if the points in the same pair have cosine similarity higher than dissimilarity threshold $\epsilon$. In this implementation, we set $\epsilon$ to 0.1 and discard one of the points within the data pair considered as a semantic duplicate.

After applying MinHash and SemDeDup, we manually review each dataset to eliminate recurring text templates. Owing to computational constraints, we further conducted additional sampling of the OSCAR dataset and the Common Crawl dataset to mitigate the issue related to dataset size. In the long run, there are around 13 billion tokens left in the corpus. The detailed information of resulting training dataset is reported in Table ~\ref{table:PTdatadistribution}.

\begin{table}
\centering
\begin{tabular}{lccc}
\toprule
\multicolumn{3}{c}{Entire dataset size $\approx$ 13 billion tokens} \\
\midrule
Data source & Proportion(\%) & Source Lang.\\
\midrule
Zhihu-KOL & 4.96 & SC \\
Wikipedia-zht & 17.03 & TC  \\
OSCAR 23.01 (zh subset) & 9.04 & SC \\
YeungNLPfirefly-train-1.1M & 2.00 & SC \\
RefinedWeb & 20.53 & EN \\
Common Crawl & 34.6 & SC+TC \\
Books & 0.83 & SC+TC \\
WebDocs & 11.01 & SC+TC \\
\bottomrule
\end{tabular}
\caption{Fraction and source language of data from each source in the training dataset. We use the following abbreviations to denote different languages. \textbf{SC}: Simplified Chinese. \textbf{TC}: Traditional Chinese. \textbf{EN}: English.}
\label{table:PTdatadistribution}
\end{table}

\subsection{Continual Pre-training Objective}
In the continual pre-training (CPT) stage, we further pretrain Bailong models with the conventional Casual Language Modeling (CLM) task. Given a sequence of input tokens $x=\{x_{1},...,x_{n}\}$, the model is trained by using an autoregressive approach to predict the next token based on preceding tokens. Within this framework, the training objective is to minimize the negative log-likelihood, which could be mathematically expressed as:
\begin{equation}
    \mathcal{L}_{\text{CLM}}(\Theta) = \E_{x\sim D_{\text{CPT}}}\,\bigg[-\sum_{i}\text{log}\,p(x_{i}\mid x_{1},...,x_{i-1};\Theta)\bigg],
\end{equation}
where $\Theta$ denotes the model parameters and $D_{\text{CPT}}$ denotes the dataset of continual pre-training.

\subsection{Training Detail}
For continuous pre-training, we conduct a single epoch of training using a paged 8-bit AdamW optimizer and a batch size of 64. We tie the embedding layer and LM head, initializing the embeddings of newly introduced tokens through zip-tie embedding initialization. LoRA modules are added to embedding layer, LM head, and all other linear layers of the base model. We set LoRA $r=64, \alpha=16$ and LoRA dropout to 0.1. QLoRA and mixed-precision training techniques are applied. With this setup, the number of trainable parameters accounts for only $4.6\%$ of the total, significantly reducing computational demands during training. The peak learning rate is set to $3e-4$ with cosine learning rate schedule, such that the final learning rate is $10\%$ of the peak learning rate value. We use a weight decay of 0.1, warmup steps of $2,000$, and gradient clipping of 1.0 to ensure the stability of training. The context length is set to be 2048, and all samples exceeding 2048 tokens are truncated. We refer to the model trained at this stage as Bailong 7B.

\section{Supervised Fine-tuning}
Although pre-trained LLMs have exhibited impressive performance across various NLP tasks, they frequently encounter challenges in accurately following human intent \citep{ouyang2022training}. To overcome this problem, supervised fine-tuning \citep{wei2021finetuned} and reward-based methods \citep{ouyang2022training, dong2023raft, rafailov2023direct, ethayarajh2024kto, hong2024orpo} are proposed to align model outputs with user preferences. In this work, we focus on enhancing model capability of following instruction by supervised fine-tuning (SFT) and leave the reward-based methods to future work. 

\subsection{Supervised Fine-tuning Method}
After continual pre-training stage, we train Bailong 7B using an instruction-following dataset. To improve the model performance in multi-turn dialogue use cases, each data in the dataset consists of one or more instructions and the corresponding outputs concatenated with end-of-sequence (EOS) tokens. Specifically saying, given a set of instructions $Instruction = \{instruction_{1},...,instruction_{n}\}$ and its output counterpart $Output = \{output_{1},...,output_{n}\}$, the template used during supervised fine-tuning is designed as:

\begin{equation}
\label{eqn:prompt template}
\text{<s>}instruction_{1}\text{</s>}output_{1}\text{</s>}...\text{</s>}instruction_{n}\text{</s>}output_{n}\text{</s>},
\end{equation}

where <s> is the BOS token and </s> is the EOS token. Similar to continual pre-training stage, we leverage the autoregressive objective during SFT stage but the loss is only computed on the output part of the template. Given a sequence of input tokens $x=\{x_{1},...,x_{n}\}$, the loss during SFT stage can then be expressed as:
\begin{equation}
    \mathcal{L}_{\text{SFT}}(\Theta) = \E_{x\sim D_{\text{SFT}}}\,\bigg[-\sum_{x_{i}\in Output}\text{log}\,p(x_{i}\mid x_{1},...,x_{i-1};\Theta)\bigg],
\end{equation}
where $\Theta$ denotes the model parameters and $D_{\text{SFT}}$ denotes the dataset of supervised fine-tuning.

\subsection{Instruction Dataset}
Generating an instruction dataset consisting of complex natural language instructions across a wide range is crucial for aligning LLMs with human preference. A conventional approach to accomplish this involves engaging human labelers to furnish their favored outputs corresponding to given prompts \citep{ouyang2022training, kopf2024openassistant}. Nevertheless, collecting data in such a manner is costly and time-consuming, posing significant challenges to scalability and generality in the curation process of training dataset. Alternative methods leverage state-of-the-art proprietary teacher LLMs (e.g. ChatGPT and GPT4) to generate target output for given prompts, enabling the cost-effective production of large-scale synthetic datasets for training. The instructions in these methods can either be manually collected or expanded by LLMs based on a small number of human-written seeds \citep{wang2022self, xu2023baize}.

Table \ref{table:SFTdatadistribution} showcases an overview of our training dataset at SFT stage. Our training dataset comprises datasets generated through various aforementioned methodologies. We briefly introduce the creation process of each dataset here: Alpaca-gpt4-zh \citep{peng2023instruction} and Moss-003-sft-data are generated by self-instruct \citep{wang2022self} method, where we sample 50,000 data from Moss-003-sft-data; ShareGPT-CN is a translated ChatGPT conversation dataset collected from ShareGPT\footnote{\url{https://sharegpt.com/}}; the contents of DRCD \citep{shao2018drcd}, FGC, and Medical are gathered from publicly available sources; GSM8k \citep{cobbe2021training} is a human-crafted dataset comprising 8.5k grade school math word problems; finally, given previous research suggesting that the self-instruction method struggles to generate complex instructions \citep{cui2023ada}, we introduce INNX, a dataset manually crafted by ourselves. INNX contains data with longer instruction lengths to augment the overall complexity of the training dataset instructions. The pipeline of dataset curation comprises three steps: 1) translation, 2) instances generation using LLMs, 3) concatenating single-turn dialogues into a multi-turn dialogue data. 

\begin{table}
\centering
\begin{threeparttable}
\resizebox{\textwidth}{!}{%
\begin{tabular}{lccccc}
\toprule
\multicolumn{6}{c}{Entire dataset size $\approx$ 120K data} \\
\midrule
Data source & Avg Input Len. & Avg Output Len. & Proportion(\%) & Avg Turns & Source Lang.\\
\midrule
ShareGPT-CN\tnote{1} & 50.76 & 275.87 & 25.96 & 3.96 & SC \\
Alpaca-gpt4-zh \citep{peng2023instruction}\tnote{2} & 27.86 & 329.72 & 16.30 & 2.50 & SC  \\
Moss-003-sft-data\tnote{3} & 49.37 & 636.49 & 45.85 & 5.44 & SC+EN \\
Medical\tnote{4} & 81.59 & 198.67 & 1.50 & 4.0 & SC \\
FGC \citep{CiteFGC2020} & 215.09 & 144.41 & 0.08 & 3.0 & TC \\
DRCD \citep{shao2018drcd} & 186.22 & 40.64 & 1.64 & 3.0 & TC \\
GSM8k \citep{cobbe2021training}\tnote{5} & 87.34 & 106.38 & 1.56 & 4.0 & EN \\
INNX (closed-source) & 253.45 & 192.12 & 7.10 & 3.34 & TC \\
\midrule
Total & 61.47 & 477.27 & 100 & 4.34 & SC+TC+EN\\
\bottomrule
\end{tabular}}
\begin{tablenotes}
     \item[1] \url{https://huggingface.co/datasets/FreedomIntelligence/ShareGPT-CN}
     \item[2] \url{https://github.com/Instruction-Tuning-with-GPT-4/GPT-4-LLM/tree/main}
     \item[3] \url{https://huggingface.co/datasets/YeungNLP/moss-003-sft-data}
     \item[4] \url{https://huggingface.co/datasets/shibing624/medical}
     \item[5] \url{https://github.com/openai/grade-school-math}
   \end{tablenotes}
\end{threeparttable}
\caption{Detailed information and source of instances in the instruction-tuning dataset. We use the following abbreviations to denote different languages. \textbf{SC}: Simplified Chinese. \textbf{TC}: Traditional Chinese. \textbf{EN}: English.}
\label{table:SFTdatadistribution}
\end{table}

\paragraph{Language translation}
For instruction-following datasets originally consisting of Simplified Chinese (ShareGPT-CN, Alpaca-gpt4-zh, Moss-003-sft-data, and Medical), we translate them into Traditional Chinese using the OpenCC library. For training set of GSM8k, we translate it from English into Traditional Chinese using ChatGPT (GPT-3.5-turbo) API. Note that the English part of Moss-003-sft-data is unaltered in this process in order to mitigate the potential catastrophic forgetting during training.

\paragraph{Generating instances utilizing LLMs}
Each data in DRCD and FGC includes a paragraph and the associated questions. To format instructions based on these data, we manually create instruction templates and fill them with paragraphs and questions of the datasets. We then utilize GPT-3.5-turbo to produce output answers for these instruction instances, thereby generating synthetic datasets composed of single-turn dialogue data.

\paragraph{Concatenating single-turn dialogues into multi-turn dialogues}
For single-turn dialogue datasets (Medical, FGC, DRCD, and GSM8k) that have single (instruction , output) pair in each instances, multiple instances are transformed into one multi-turn dialogue instance by simply concatenating them. The instances are subsequently separated by EOS token as the format of \eqref{eqn:prompt template} during training.

\subsection{Fine-tuning Detail}
During SFT stage, we train the model for five epochs with paged 8-bit AdamW optimizer and a batch size of 64. We implement mix-precision training and QLoRA to reduce the memory usage of training. The LoRA modules are inserted to all linear layers of the pre-trained model, with LoRA $r=64, \alpha=16$ and LoRA dropout of 0.05. The peak learning rate is set to $1e-4$ with cosine learning rate schedule. Moreover, we use warmup ratio of $0.03$ and max norm of the gradients of 0.3 to ensure the stability of training. Finally, the context length is set to be 2048, and any samples exceeding 2048 tokens are truncated. Under this configuration, the number of trainable parameters accounts for $4.12\%$ of the total. The resulting model trained at this stage is called Bailong-instruct 7B. 

\section{Evaluation}

\subsection{Bailong-Bench}
Despite the existence of numerous Traditional Chinese benchmarks tailored for language model evaluation, inherent limitations emerge when applying them in conjunction with traditional metrics to assess modern performant LLMs. These limitations primarily stem from the difficulty of using traditional metrics to gauge semantic understanding of natural language and the narrow scope of benchmarks coverage. To overcome these limitations, there is a surge in demand for developing more diverse and complex Traditional Chinese benchmarks to ensure comprehensive and accurate evaluations of the capabilities of LLM.

To solve this issue, we propose Bailong-bench, a benchmark dataset comprising 140 instructions written in English and Traditional Chinese. The main purpose of Bailong-bench is to evaluate the cpabilities in following instructions, detecting harmful inputs, and engaging in multi-turn dialogue. Specifically, the benchmark is composed of 14 categories containing creative writing, mail assistant, health consultant, translation, copywriting generation, knowledge-based question, summarization, proofreading, open question, morality and ethics, general question, English instruction, arithmetic, and multi-turn dialogue. Each category has 10 meticulously designed instructions. Table \ref{table:Bailong-bench sample} shows several sample instructions in Bailong-bench.

\begin{table}[ht]
\centering
\resizebox{\textwidth}{!}{%
\begin{tabular}[c]{c|cp{10cm}}
\toprule
Category & Turns & Instruction sample\\
\midrule
Creative writing & 1 & 我是一家公司的老闆，如今公司正面臨著財務危機，我需要發表演講以激勵員工，幫我寫一封演講稿。\\
\midrule
Arithmetic & 1 & 班上有30位學生，老師想要給每位學生一本數學書和一枝鉛筆。數學書每本15元，鉛筆每枝5元。老師總共需要支付多少錢？\\
\midrule
Translation & 1 & 翻譯以下文章成繁體中文\textbackslash nIn a world driven by economic complexities, the importance of financial literacy cannot be overstated...\\
\midrule
English instruction & 1 & What are the pros and cons of high interest rates?\\
\midrule
\multirow{3}{*}{Multi-turn} & 1 & 將以下句子裡的'a'全都改寫成'A'\textbackslash n A cat is walking on the street\\
& 2 & 再幫我把句子裡的't'改成'T'\\
& 3 & 把句子還原成最初的大小寫形式\\
\bottomrule
\end{tabular}}
\caption{Sample instructions in Bailong-bench.}
\label{table:Bailong-bench sample}
\end{table}

\subsection{Evaluation Method}
We implement single answer grading and reference-guided grading variants of LLM-as-a-judge \citep{zheng2024judging} to evaluate model outputs owing to their effectiveness and scalability. During the evaluation, we prompt GPT-4 turbo to serve as a judge and assign it to score the model response to given testing instruction. The benchmarks are categorized as follows:
\paragraph{Reading Comprehension:}We use the test sets of FGC and DRCD to measure the model performance on contextual question answering. Given a paragraph and question from FGC and DRCD, we combine them with a preface to form a prompt as an input to the model. 
\paragraph{Sentiment Analysis:}We make use of the translated IMDB (IMDB-TC) dataset \citep{maas-EtAl:2011:ACL-HLT2011} to test the ability of sentiment analysis. Given a movie review from the dataset, we prompt model to determine the emotion of the paragraph. 
\paragraph{World Knowledge:}We use TTQA dataset to test the model's familiarity with Taiwanese culture.
\paragraph{Summarization:}Translated Xsum (Xsum-TC) dataset \citep{Narayan2018DontGM} is used to evaluate the model's capability of text summarization. Given a document in the dataset, the model is requested to generate the corresponding Summary for the document.
\paragraph{Comprehensive Benchmark:}We make use of MT-bench and Bailong-bench to comprehensively evaluate the performance of models across diverse domains.

After the generated responses are filled into the evaluation prompt templates, the judge model scores the responses from zero to ten based on several factors including helpfulness, relevance, accuracy, depth, creativity, and level of detail of the response. Furthermore, the judge is required to score zero if the language of model response differs from the language of the instruction. We present the prompt templates in Appendix \ref{prompt templates}. For the judgement of FGC, IMDB-TC, DRCD, and Xsum-TC, we leverage Traditional Chinese single answer grading prompt template shown in Figure \ref{fig:prompt template single score grading}. For TTQA, we use reference-guided grading template presented in Figure \ref{fig:prompt template single score ref}. For MT-bench, we use English multi-turn answer grading prompt template (shown in Figure \ref{fig:prompt template MT-bench}). Finally, for Bailong-bench, Traditional Chinese multi-turn prompt template (shown in Figure \ref{fig:prompt template multi-turn}) is adopted for judging the multi-turn dialogue category of Bailong-bench and Traditional Chinese single answer grading prompt template is used for the other categories. 
\subsection{Evaluation Results}
We present the evaluation results of our proposed Bailong-instruct-7B and other LLMs that support Traditional Chinese in Table \ref{table:Evaluation results}. As Table \ref{table:Evaluation results} shown, Bailong-instruct-7B comprehensively outperforms Llama-2-13B-chat \citep{touvron2023llama}, Llama-2-7B-chat \citep{touvron2023llama}, and Gemma-7B-it \citep{team2024gemma} in all of Traditional Chinese benchmarks. Thorough examination, we find that although Llama-2-7B-chat and Llama-2-13B-chat possess a certain level of understanding of Traditional Chinese, they tend to generate English responses after receiving Traditional Chinese instructions. Such behavior significantly misaligns with the preferences of Traditional Chinese users. Table \ref{table:example one} showcases an comparing example between Bailong-instruct-7B and Llama-2 chat. Meanwhile, Bailong-instruct-7B also outperforms Taiwan-LLM-7B-v2.1-chat \citep{lin2023taiwan} in 5 out of 7 evaluation benchmarks and surpasses the overall performance of Taiwan-LLM-13B-v2.0-chat \citep{lin2023taiwan} with fewer parameters. Overall, the results have demonstrated the efficiency and effectiveness of our training framework. It is worth noting that Bailong-instruct-7B performs worse on TTQA dataset compared to Taiwan-LLM-7B-v2.1-chat and Taiwan-LLM-13B-v2.0-chat. We speculate that this discrepancy may be due to Taiwan-LLM's deliberate reinforcement of the model understanding of Taiwanese culture and linguistic nuances during the training phase. Future work may focus on improving this aspect to align the model's outputs more closely with the preferences of Taiwanese users.

\begin{table}[h]
\resizebox{\textwidth}{!}{%
\centering
\begin{tabular}[c]{l|cccccccc}
\toprule
Model & FGC & IMDB-TC & DRCD & Xsum-TC & TTQA & MT-bench & Bailong-bench & Avg\\
\midrule
GPT-3.5-turbo & 7.36 & 6.35 & 7.23 & 6.52 & 4.97 & 7.65 & 9.39 & 7.07\\
\textbf{Bailong-instruct-7B} & 6.66 & 5.53 & 5.56 & 5.22 & 1.98 & 2.32 & 9.35 & 5.23\\
Taiwan-LLM-13B-v2.0-chat & 5.92 & 4.35 & 5.33 & 5.51 & 4.69 & 1.95 & 7.89 & 5.09\\
Taiwan-LLM-7B-v2.1-chat & 4.46 & 2.04 & 4.02 & 4.95 & 2.49 & 2.36 & 6.71 & 3.86\\
Llama-2-13b-chat & 1.50 & 1.88 & 1.11 & 3.45 & 0.37 & 5.18 & 2.89 & 2.34\\
Llama-2-7b-chat & 1.38 & 2.33 & 1.10 & 2.80 & 0.53 & 4.75 & 2.64 & 2.22\\
Gemma-7B-it & 1.32 & 0.07 & 0.50 & 0.71 & 0.77 & 1.06 & 4.46 & 1.27\\

\bottomrule
\end{tabular}}
\caption{Instruction-following evaluation results judged by GPT-4-turbo. Given an instruction and the corresponding model output, the LLM judge is prompted to score the output based on several factors such as accuracy, level of detail, language consistency, and so forth.}
\label{table:Evaluation results}
\end{table}

\section{Conclusion}
Our work on Bailong has demonstrated a memory-efficient training framework to adapt English-dominated open-source LLMs to Traditional Chinese by leveraging parameter-efficient tuning methods and advanced embedding initialization. To comprehensively evaluate the fine-tuned model performance in various real-world applications, we also propose Bailong-bench benchmark dataset. The evaluation results on Bailong-bench and other benchmarks show that our instruction-tuned model not only improves Traditional Chinese understanding and text generation capability of its predecessor Llama 2, but also achieves competitive Traditional Chinese performance compared to other models trained in a full-parameter tuning fashion. Furthermore, our proposed training framework in principle can generalize to the adaption of any other low-resource languages, we believe that it can benefit the research community and facilitate the democratisation of large language model development. Future work may scale up the training dataset to further refine the model performance on Traditional Chinese tasks. Also, future work will implement preference tuning (e.g. RLHF and DPO) for superior alignment with user intents.

\section*{Limitations}
While Bailong has demonstrated remarkable Traditional Chinese understanding, generating, and instruction-following capabilities, the model still has several limitations, including: 
\begin{itemize}
\item Hallucination: As with all LLMs, the model may generate responses that are factually incorrect, illogical, or irrelevant to the given inputs.

\item Language limitation: The model is mainly trained to follow Traditional Chinese and English instructions. Therefore, the model may lack the capability to follow instructions composed in other languages including Simplified Chinese, leading to potential misinterpretations or errors in response. 

\item Potential toxicity and uncontrollability: Due to the lack of further training through reinforcement learning from human feedback, there is a risk that model may produce harmful responses or deviate from the expected responses.

\item Biases from training data: Since LLMs are trained on large scale of training data from various sources, they may inherit social biases and misconceptions existing in the data. This can result in the generation of biased outputs, thereby potentially causing negative impacts on both users and society at large.
\end{itemize}

\subsubsection*{Acknowledgments}
We express our sincere gratitude to the open-source community for their support during the active phase of our research. Special thanks go to Chao Chuan Ko, Chih Cheng Chang, and  Pei Yu Lee for their help to our project. Additionally, we would like to convey our heartfelt thanks to Ban Xi Lin for insightful discussions regarding the zip-tie embedding initialization.

\bibliography{iclr2021_conference}
\bibliographystyle{iclr2021_conference}
\clearpage
\appendix
\section{Prompt Templates} \label{prompt templates}
We present the prompt templates we used when implementing LLM-as-a-judge method to evaluate model performance as follow.
\begin{figure}[h]
  \centering
  \includegraphics[width=\linewidth]{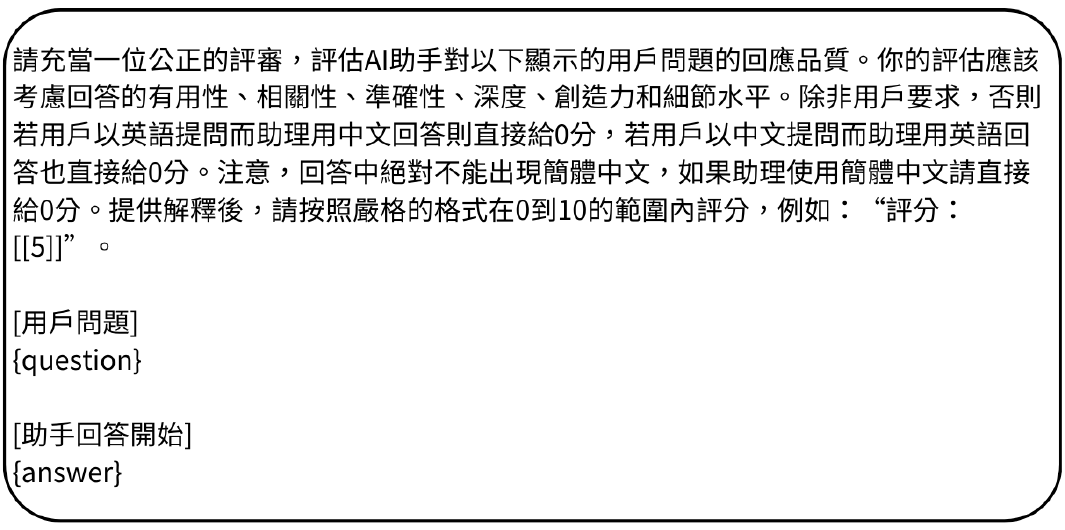}
  \caption{The prompt template for single-turn answer grading.}
  \label{fig:prompt template single score grading}
\end{figure}

\begin{figure}[h]
  \centering
  \includegraphics[width=\linewidth]{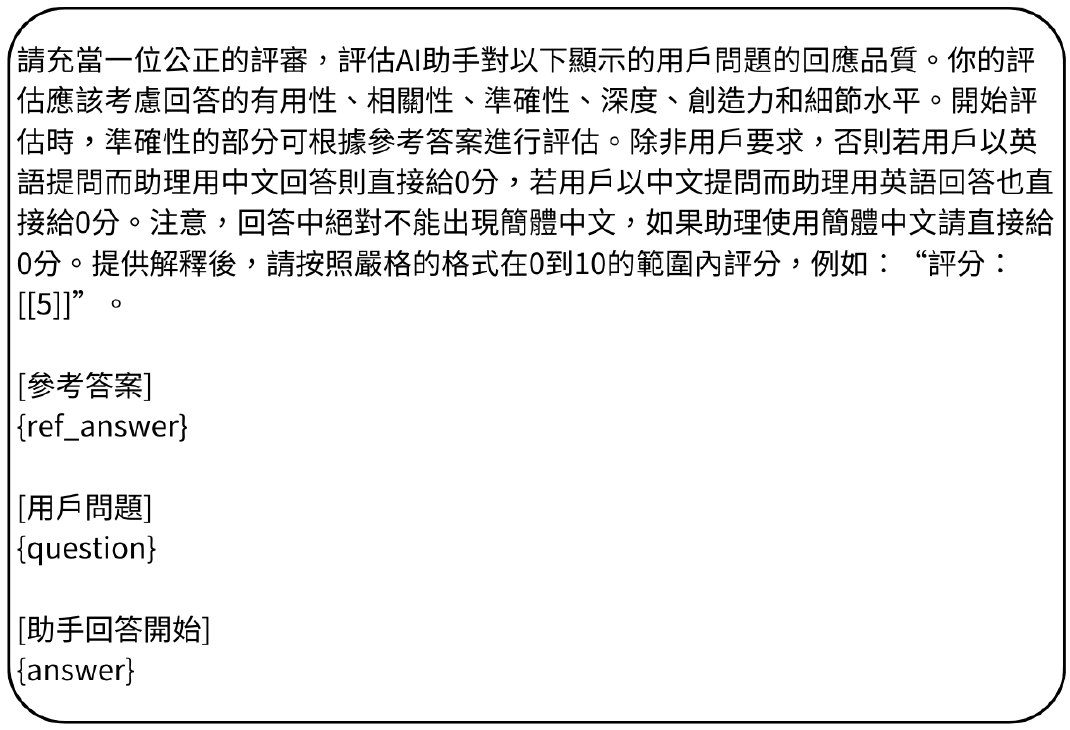}
  \caption{The prompt template for single-turn answer grading provided with reference answer.}
  \label{fig:prompt template single score ref}
\end{figure}

\begin{figure}[h]
  \centering
  \includegraphics[width=\linewidth]{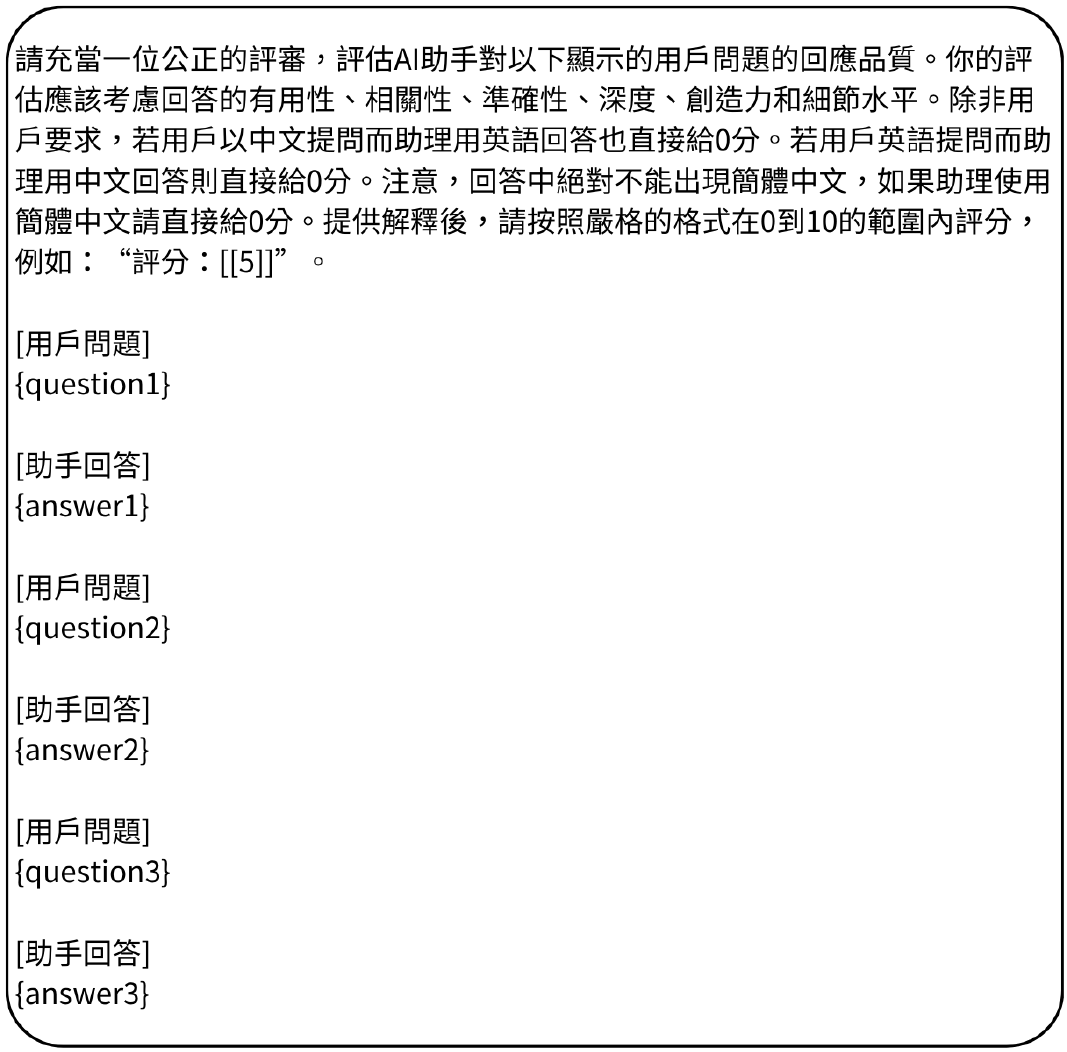}
  \caption{The prompt template for multi-turn answer grading.}
  \label{fig:prompt template multi-turn}
\end{figure}

\begin{figure}[h]
  \centering
  \includegraphics[width=\linewidth]{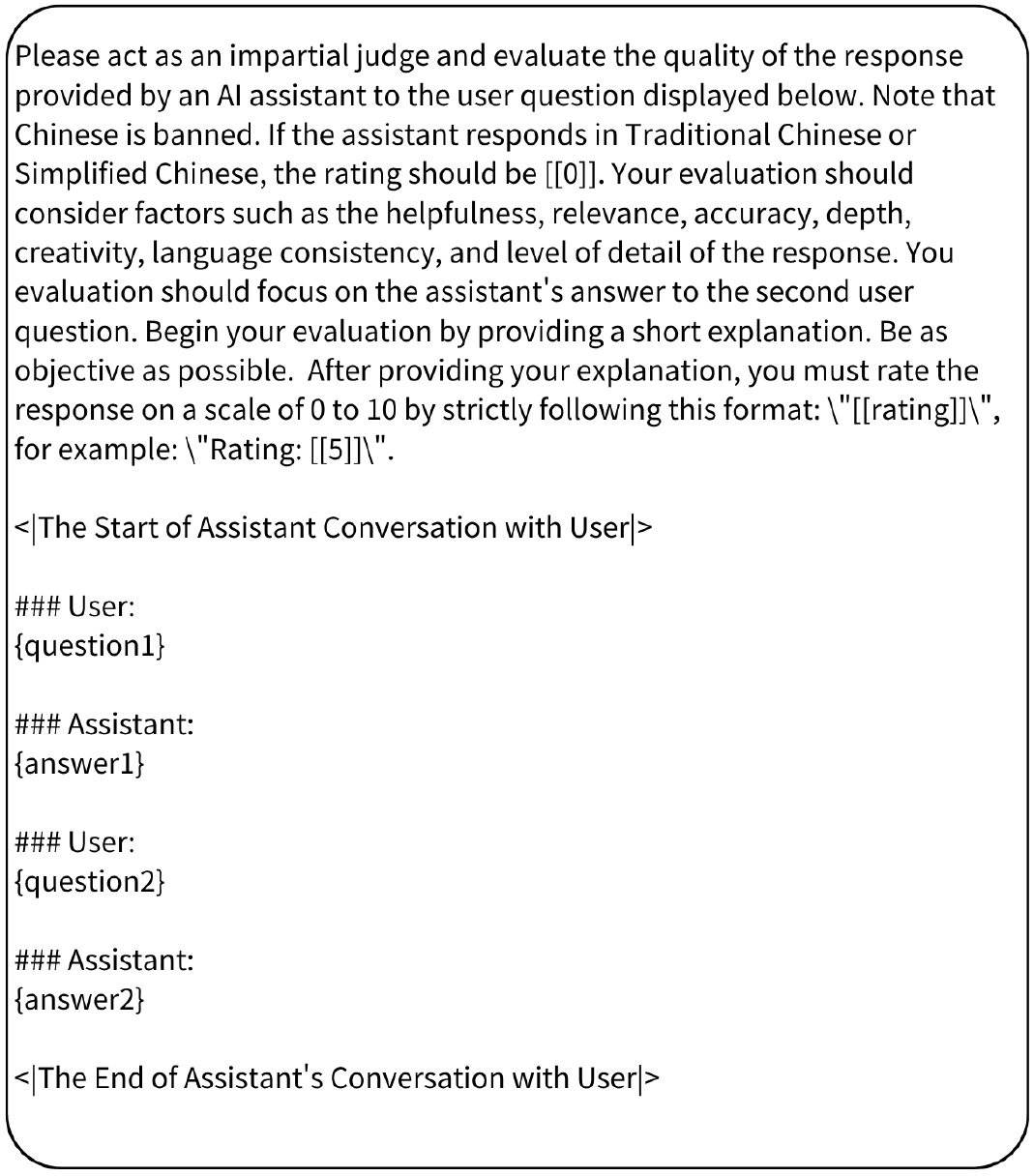}
  \caption{The prompt template for MT-bench.}
  \label{fig:prompt template MT-bench}
\end{figure}

\clearpage

\section{Comparison Example between Bailong-instruct-7B and Llama-2 Chat Series}

\begin{table}[ht]
\centering
\resizebox{\textwidth}{!}{%
\begin{tabular}{P{3cm}|p{13cm}}
\toprule
Model & Instruction: 寫一封信給我的上司，告訴他我今天頭痛，想請病假。\\
\midrule
Bailong-instruct-7B\newline(10/10) & 親愛的[上司]，\newline\newline希望你一切安好。我寫信是告訴你我今天頭痛，需要請病假。我已經進行了檢查，醫生建議休息和恢復精神狀態。我會盡快返回工作，並隨時提供任何支援。\newline\newline感謝您的理解，並期待儘快回到工作崗位。\newline\newline此致，\newline[你的名字]\\
\midrule
Llama-2-7B-chat\newline(0/10) & Dear [Supervisor's Name],\newline\newline I hope this email finds you well. Unfortunately, I am not feeling too great today and I am writing to request a sick leave. I am experiencing a severe headache and I believe it is best if I take a day off to rest and recover.\newline\newline I understand that this may be an inconvenient time, but I assure you that I will make up for the lost time as soon as possible. I am committed to my work and I will do everything in my power to ensure that my absence does not negatively impact the team.\newline\newline Please let me know if there are any further steps I need to take or if there are any alternative arrangements that can be made. Thank you for your understanding and support.\newline\newline Sincerely,\newline[Your Name]\\
\midrule
Llama-2-13B-chat\newline(0/10) & Dear [上司的名字],\newline\newline I hope this message finds you well. Unfortunately, I am not feeling well today and am experiencing a severe headache. I would like to request a sick leave for today to take care of my health.\newline\newline I apologize for any inconvenience this may cause and will make sure to catch up on any missed work as soon as I am feeling better. Please let me know if there are any issues or if there is anything I can do to make the process smoother.\newline\newline Thank you for your understanding and support.\newline\newline Sincerely,\newline[Your Name]\\
\bottomrule
\end{tabular}}
\caption{Example of outputs from Bailong-instruct-7B and Llama-2 chat models}
\label{table:example one}
\end{table}

\end{CJK*}
\end{document}

%% file: math_commands.tex
\usepackage{amsmath,amsfonts,bm}

\def\eqref#1{equation~\ref{#1}}

\def\1{\bm{1}}

\DeclareMathAlphabet{\mathsfit}{\encodingdefault}{\sfdefault}{m}{sl}
\SetMathAlphabet{\mathsfit}{bold}{\encodingdefault}{\sfdefault}{bx}{n}

\newcommand{\E}{\mathbb{E}}

